%% file: main.tex
\documentclass[letterpaper,10pt,conference]{ieeeconf}
\IEEEoverridecommandlockouts
\usepackage{lipsum}
\usepackage{colortbl}
\usepackage{url}
\usepackage{graphicx}
\usepackage{amsfonts}
\usepackage{soul} %
\usepackage{color, xcolor} %
\usepackage{amsmath}
\usepackage{amssymb}
\usepackage{multirow}
\usepackage{array}
\usepackage[linesnumbered,ruled,vlined]{algorithm2e}
\usepackage{booktabs}
\usepackage{placeins}
\usepackage[noadjust]{cite}
\usepackage{soul}
\usepackage[colorlinks,allcolors=black]{hyperref}
\usepackage{color}

\usepackage{capt-of}
\newsavebox{\experimentrewardbox}
\usepackage{threeparttable}
\usepackage{arydshln}

\begin{document}

\title{\LARGE\bfseries FlashNav: Training Deployable Robot Navigation Policies in Seconds}

\author{Shanze~Wang$^{1,2,*}$, Yiwei~Qian$^{1,3,*}$, Xinming~Zhang$^{1,4}$, Jun~Xue$^{1,5}$,\\
Siwei~Cheng$^{1,4}$, Xianghui~Wang$^{1,6}$, Qingyuan~Hu$^{1,4}$, Yanjun~Chen$^{1,7}$,\\
Xiaoyu~Shen$^{1}$, Hailong~Huang$^{2}$, and Wei~Zhang$^{1,\dagger}$%
\thanks{$^{*}$The first two authors contributed equally to this work.}%
\thanks{$^{1}$Eastern Institute of Technology, Ningbo, China.}%
\thanks{$^{2}$Department of Aeronautical and Aviation Engineering, The Hong Kong Polytechnic University, Hung Hom, Kowloon, Hong Kong.}%
\thanks{$^{3}$National University of Singapore, Singapore.}%
\thanks{$^{4}$University of Science and Technology of China, Hefei, China.}%
\thanks{$^{5}$Shanghai Jiao Tong University, Shanghai, China.}%
\thanks{$^{6}$Department of Mechanical Engineering, The Hong Kong Polytechnic University, Hung Hom, Kowloon, Hong Kong.}%
\thanks{$^{7}$Department of Computing, The Hong Kong Polytechnic University, Hung Hom, Kowloon, Hong Kong.}%
\thanks{$^{\dagger}$Wei Zhang is the corresponding author: \texttt{zhw@eitech.edu.cn}.}%
}
\hypersetup{
  pdftitle={FlashNav: Training Deployable Robot Navigation Policies in Seconds},
  pdfauthor={Shanze Wang, Yiwei Qian, Xinming Zhang, Jun Xue, Siwei Cheng, Xianghui Wang, Qingyuan Hu, Yanjun Chen, Xiaoyu Shen, Hailong Huang, Wei Zhang}
}

\maketitle
\pagestyle{empty}  %
\thispagestyle{empty} %
\begin{abstract}
Training Deep Reinforcement Learning (DRL) navigation policies for different robot configurations remains time-consuming. We present FlashNav, a GPU-based framework that trains robot-specific navigation policies within tens of seconds. A unified robot specification configures a lightweight simulator for batched motion updates, range sensing, and footprint collision checking over a shared occupancy map. The framework supports nonconvex footprints, different sensor configurations and drive types. Blockwise ray queries and selective observation recomputation after resets reduce simulation overhead, while GPU-resident replay and overlapping experience collection and learner updates support efficient off-policy training. Experiments covered five robot configurations and three computing platforms. With FastDSAC on a single RTX~5090 GPU, FlashNav can train a deployable navigation policy in under 30 seconds. FlashNav achieved the highest success rate and score in the benchmark comparison. The selected policies were deployed on wheeled, quadrupedal, humanoid, and irregularly shaped robots without additional policy training.
\end{abstract}

\begin{keywords}
Robot navigation, deep reinforcement learning, GPU simulation, sim-to-real transfer.
\end{keywords}

\suppressfloats[t]
\begin{figure}[t]
\centering
\includegraphics[width=\columnwidth]{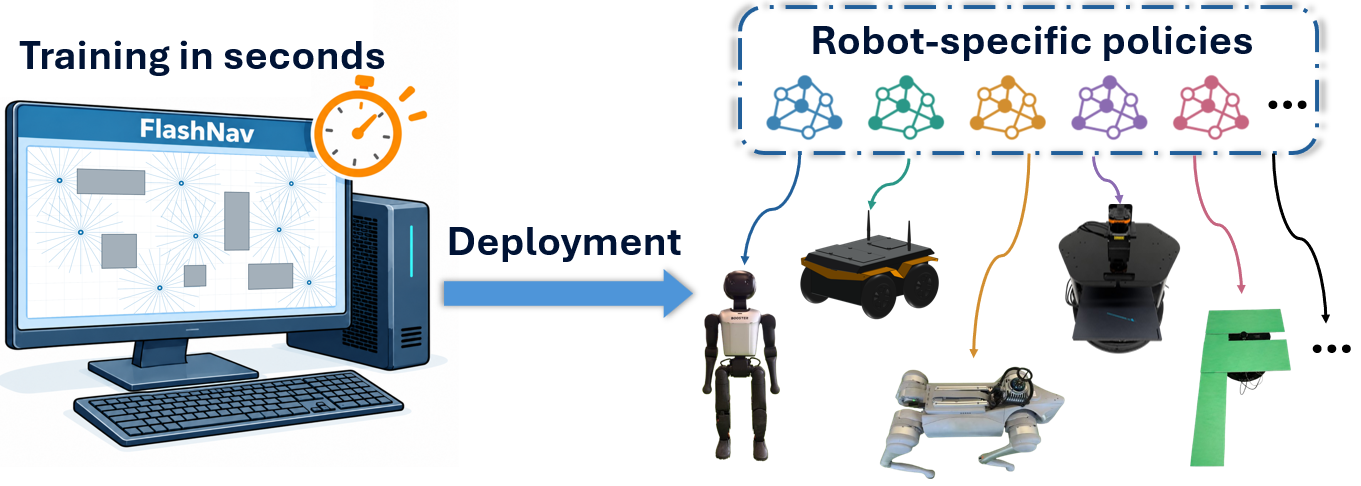}
\caption{FlashNav trains robot-specific navigation policies within seconds using GPU-based simulation. Robot specifications support different footprints, sensors, and motion interfaces.}
\label{fig:flashnav_overview}
\end{figure}

\section{Introduction}
\label{sec:introduction}

DRL-based local planners learn to map range observations and goal information to continuous velocity commands, supporting navigation in cluttered environments~\cite{Zhang2022IPAPRec,xie2023drlvo}. Recent methods also improve collision avoidance and interaction modeling in dynamic or crowded scenes~\cite{xu2025navrl,liu2026height}, while model-based learning supports navigation across different robots~\cite{han2025neupan}. However, training navigation policies in simulation can still take hours or even days~\cite{pfeiffer2018reinforced,akmandor2022deep}. This cost slows algorithm iteration, platform adaptation, and sim-to-real validation. Reducing training time while retaining deployable navigation behavior is therefore a practical requirement for applying DRL to diverse ground robots.

To reduce training time, platforms such as Isaac Gym enable large-scale parallel simulation on GPUs~\cite{makoviychuk2021isaacgym}. However, navigation training in Isaac Sim can still take hours~\cite{xu2025navrl}. High-fidelity rendering and detailed dynamics remain computationally costly~\cite{shacklett2021large,truong2023rethinking}. For visual navigation, batched rendering and optimized policy computation have reduced training time to 1.5 days on a single GPU~\cite{shacklett2021large}. Simplified robot motion models reduce computation and have improved sim-to-real transfer in visual navigation by enabling more training experience and reducing overfitting to simulated dynamics~\cite{truong2023rethinking}. Color combines lightweight 2D simulation with partially decoupled data collection and policy learning to train a deployable local planner in about one hour~\cite{wang2025color}. Further reducing training time while preserving robot geometry, sensing, and motion constraints remains a challenge.

To address this challenge, we present FlashNav, a GPU-based framework that trains navigation policies within tens of seconds, as shown in Fig.~\ref{fig:flashnav_overview}. Its lightweight simulator performs batched planar motion updates, range sensing, and footprint collision checking over a shared occupancy bitmap. Experience collection and large-batch off-policy learning run on the same GPU, with their execution scheduled to overlap.
FlashNav is evaluated across five robot configurations with different footprints and sensors and on the BARN benchmark~\cite{perille2020benchmarking}. Real-world tests assess sim-to-real transfer across wheeled, quadrupedal, humanoid, and irregularly shaped robots.
Our main contributions are summarized as follows:
\begin{itemize}

\item \textbf{To the best of our knowledge, FlashNav is the first DRL-based framework to train deployable range-based local navigation policies for ground robots within tens of seconds on a single consumer-grade GPU.}

\item A unified robot specification supports arbitrary planar footprints, including nonconvex shapes, and different sensors, and different drive methods within the same training pipeline.

\item Extensive simulations, including the BARN Challenge~\cite{perille2020benchmarking}, and sim-to-real experiments across different robot platforms validate the effectiveness of the trained policies.
\end{itemize}

\section{Related Work}
\subsection{Navigation Representations and Simulation}
Range observations describe distances from a sensor to obstacles, whereas collision avoidance depends on the robot footprint and feasible motion. Encoding dimensions and motion limits in policy inputs supports transfer across rectangular robots~\cite{zhang2025drldclp}, while model-based learning connects obstacle points with robot geometry to compute collision-free motion~\cite{han2025neupan}. In dynamic scenes, pedestrian motion and interactions inform navigation decisions~\cite{xie2023drlvo,liu2026height}, and safety filters constrain learned actions~\cite{xu2025navrl}. Adaptive range preprocessing also improves learning speed by emphasizing nearby obstacles~\cite{Zhang2022IPAPRec}.

For simulation-based training, these observation and motion requirements determine what the environment must reproduce. Lightweight planar simulators generate range observations and advance robot motion without full 3D scenes~\cite{kastner2024murosim}. Configurable obstacle maps support curriculum learning in geometric navigation environments with continuous velocity commands~\cite{samvelyan2024jaxnav}. Robot configuration interfaces specify kinematics, collision geometry, and LiDAR sensing~\cite{han2026irsim}, while planar physics models also capture interactions among multiple robots~\cite{bettini2022vmas}.

\subsection{High-Throughput Reinforcement Learning}
For a given simulation model, faster experience generation requires efficient environment execution and data exchange. Thread-pool execution and batched interfaces increase CPU environment throughput~\cite{weng2022envpool}. Batching simulation and rendering shares scene storage and reduces repeated loading and synchronization costs~\cite{shacklett2021large}. When simulation and learning run on the GPU, observations and actions can pass directly between them~\cite{makoviychuk2021isaacgym,lan2022warpdrive}. Heterogeneous systems coordinate CPU simulation with GPU learning through shared buffers and overlapping execution~\cite{jia2026unilab}.

Faster collection also changes the balance between experience generation and policy updates. In off-policy RL, the number of updates per collected transition controls data reuse and learning cost. Parallel training separates data collection, policy learning, and value learning so that their execution rates can be adjusted~\cite{li2023pql}. Partially decoupled collection and learning have also been used with grid-based navigation simulation~\cite{wang2025color}. Recent off-policy methods combine parallel simulation with large-batch updates for humanoid control~\cite{seo2025fasttd3}, \cite{seo2025fastsac}, \cite{xue2026fastdsac}, while efficient dynamics simulation and curriculum learning support rapid off-policy training for quadrotor control~\cite{eschmann2024fly}. Training time therefore depends on the simulation workload, the update schedule, and the experience needed to reach the target performance.

\begin{figure*}[!t]
\vspace*{6pt}
\centering
\includegraphics[width=\textwidth]{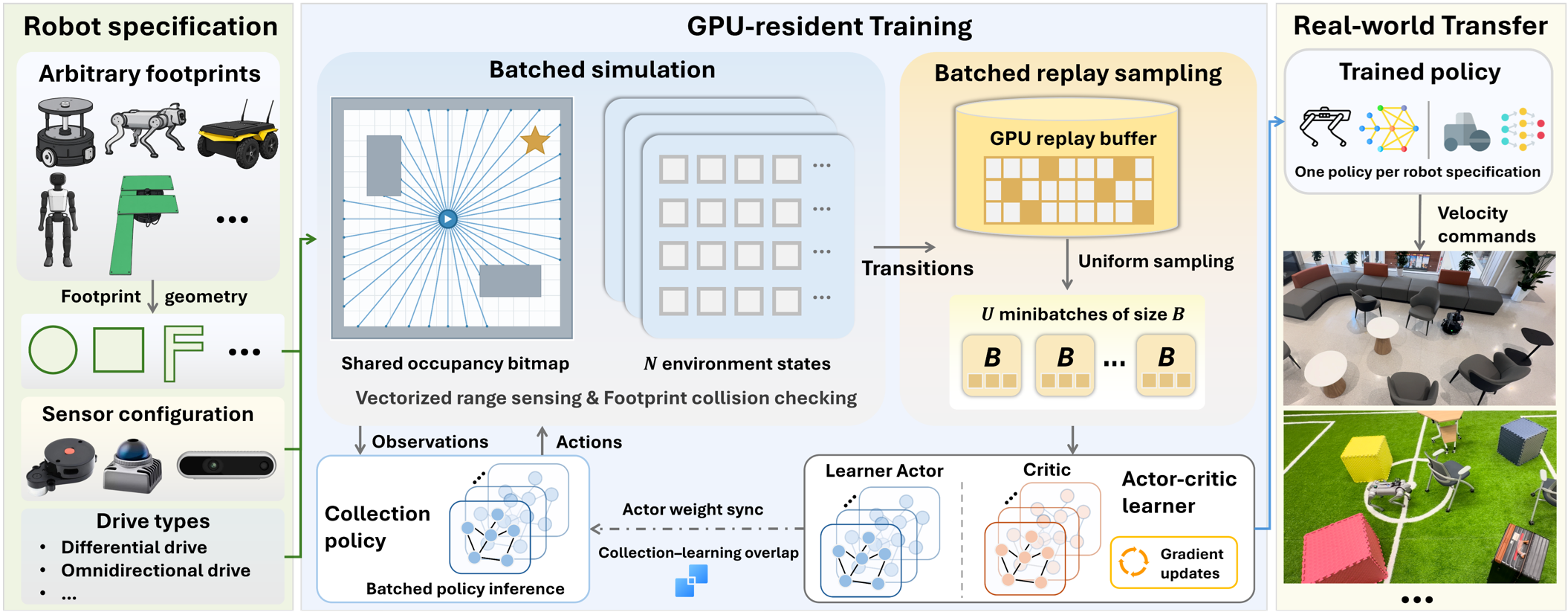}
\caption{FlashNav training framework. Robot specifications configure the batched simulator. Uniform sampling from the GPU-resident replay buffer prepares $U$ minibatches of $B$ transitions for learner updates. Updated actor weights are synchronized with the collection policy, and trained policies issue velocity commands during deployment.}
\label{fig:flashnav_framework}
\end{figure*}

\section{Preliminaries}\label{lim}
\subsection{Problem Formulation}
\label{sec:navigation_formulation}

In local navigation, a robot follows continuous velocity commands to reach a goal while avoiding obstacles. At time step $t$, its planar pose is $\mathbf q_t=(x_t,y_t,\psi_t)$, where $(x_t,y_t)$ is the position of the body-frame origin in the map frame and $\psi_t$ is the heading. A policy $\pi_\theta$, parameterized by $\theta$, maps local observations to velocity commands within the configured bounds. The commands are $(v_x,\omega)$ for differential-drive navigation and $(v_x,v_y,\omega)$ for omnidirectional navigation, where $v_x$ and $v_y$ are the longitudinal and lateral velocities in the body frame, and $\omega$ is the angular velocity. A low-level controller executes these commands. 

Each robot specification is defined as $\xi=(\mathcal F,\mathcal S,\mathcal K,\mathcal U)$, where $\mathcal F\subset\mathbb R^2$ is the occupied footprint in the body frame, $\mathcal S$ defines sensor placement and range-sensing parameters, $\mathcal K$ specifies the motion model, and $\mathcal U$ contains velocity and acceleration limits. Circular and rectangular footprints use geometric parameters, while nonconvex footprints, including the F-shaped chassis, are defined by ordered polygon vertices. Before training, the robot specification configures motion, sensing, and collision checking without requiring changes to the simulator or training pipeline.

The policy observation is $\mathbf o_t=[\mathbf z_t,d_t,\alpha_t,\mathbf v_t]$, where $\mathbf z_t$ contains sector-pooled obstacle features, $d_t$ and $\alpha_t$ denote the goal distance and direction relative to the robot, and $\mathbf v_t$ contains the current linear and angular velocities. The distance features use footprint clearance~\cite{wang2026dclppp}. Valid range endpoints are expressed in the body frame, and their shortest distances to the occupied footprint $\mathcal F$ are computed before taking the minimum within each angular sector. Following~\cite{Zhang2022IPAPRec}, the actor and critic networks transform each pooled distance $\delta$ as $\phi_\beta(\delta)=1/\max(\delta+\beta,\epsilon)$, where $\beta$ is a learned scalar offset and $\epsilon>0$ is a numerical lower bound. This transformation compresses large clearance values and increases sensitivity to small clearances, helping the robot navigate cluttered environments.
The navigation reward is based on goal progress, arrival, and collision:
\begin{equation}
\label{eq:flashnav_reward}
r_t =
\begin{cases}
r_{\text{collision}}, & \text{collision,}\\
r_{\text{success}}, & \text{goal reached without collision,}\\
c_1(d_t-d_{t+1}), & \text{otherwise,}
\end{cases}
\end{equation}
where $r_{\text{success}}=10$ rewards arrival, $r_{\text{collision}}=-10$ penalizes collision, and $c_1=2$ scales the progress reward. The policy is trained to maximize the expected discounted return.

\section{FlashNav Training Framework}
\label{sec:flashnav_framework}

Fig.~\ref{fig:flashnav_framework} summarizes the training framework. Robot specifications configure geometry, sensing, and motion for batched navigation simulation. The generated transitions pass through a replay buffer stored in GPU memory to the off-policy learner, whose actor weights update the collection policy. 

\subsection{Batched Navigation Simulation}
\label{sec:batched_simulation}

FlashNav executes navigation simulation and off-policy learning on a single GPU. The environment batch shares an occupancy bitmap, while poses, velocities, goals, and episode states are stored in tensors indexed by environment. Batched policy inference supplies velocity commands for all environments, and tensor operations advance their motion, generate observations, and compute rewards. The resulting transitions are written directly to a GPU-resident circular replay buffer.

Range sensing uses fixed-step samples along each beam. Samples from all environments are queried in distance blocks, and beams with a detected obstacle are removed from subsequent blocks. This limits temporary storage to the current block and avoids further queries beyond detected hits. When episodes end, only the corresponding environments are reset and their observations recomputed. Observations before reset are retained as next observations in replay, while reset observations are returned to the collector.

The same vectorized simulator also supports parallel policy evaluation. Each batch assigns fixed start--goal cases to separate environments and evaluates one checkpoint through deterministic batched inference. Completion masks record each case once and exclude completed cases from further accumulation of rewards and outcome statistics. Each case ends when the robot reaches the goal, collides with an obstacle, or reaches the evaluation step limit. A batch ends when all cases have ended or the evaluation step limit is reached. Additional cases are processed in subsequent batches.

\begin{figure*}[!t]
\vspace*{6pt}
\sbox{\experimentrewardbox}{%
\begin{minipage}[t]{\columnwidth}
\vspace{0pt}
\centering
\includegraphics[width=0.95\linewidth]{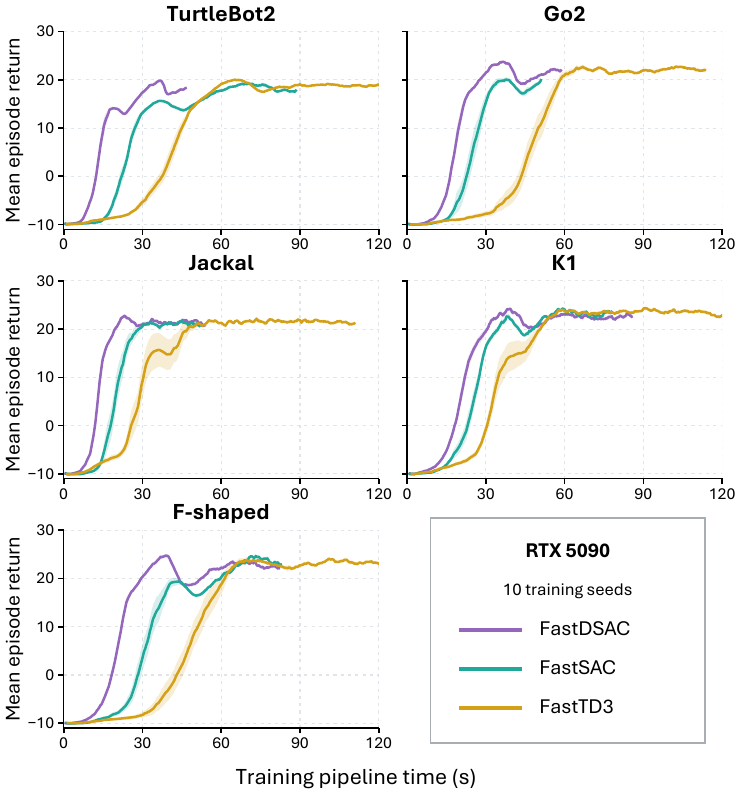}
\captionof{figure}{Mean episode returns during training on RTX 5090 across five robot configurations. Colors identify learners; curves show means across ten training seeds, with shading indicating $\pm$ standard errors.}
\label{fig:reward_time}
\end{minipage}%
}
\noindent
\begin{minipage}[t]{\columnwidth}
\vspace{0pt}

\par\vspace{1.2ex}
\noindent
\begin{minipage}[t]{0.31\linewidth}
\centering
\includegraphics[width=\linewidth]{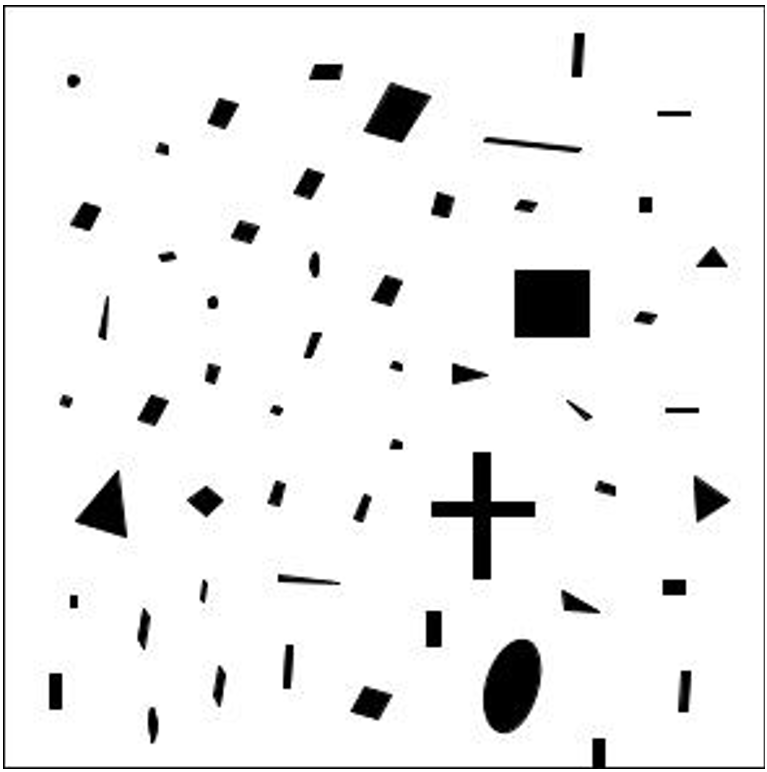}
\par\small (a) SEnv1: training
\end{minipage}\hfill
\begin{minipage}[t]{0.31\linewidth}
\centering
\includegraphics[width=\linewidth]{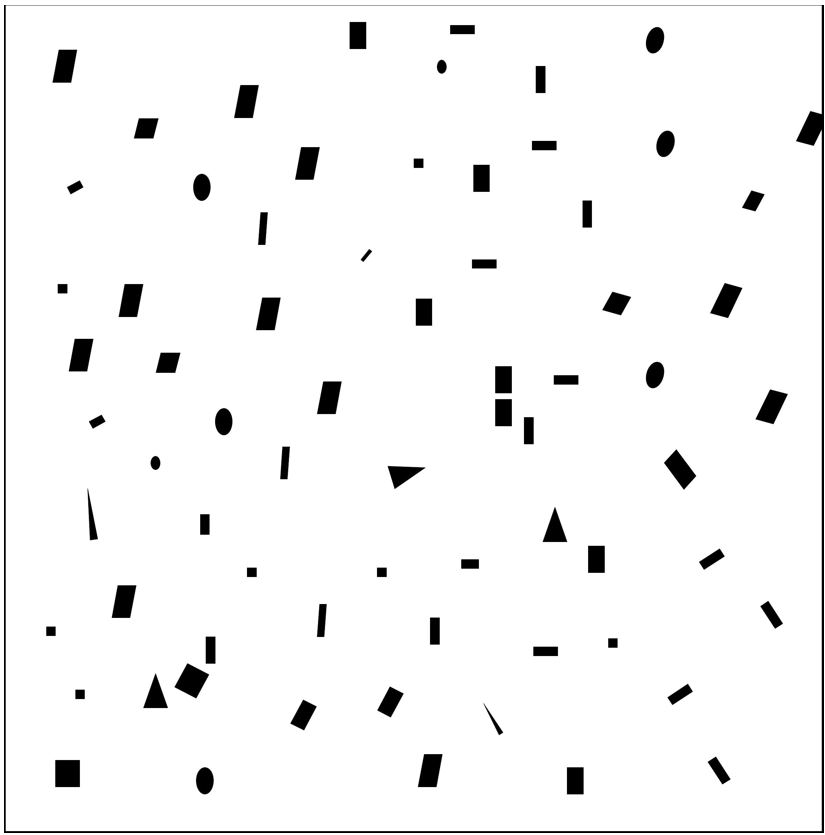}
\par\small (b) SEnv2: testing
\end{minipage}\hfill
\begin{minipage}[t]{0.305\linewidth}
\centering
\includegraphics[width=\linewidth]{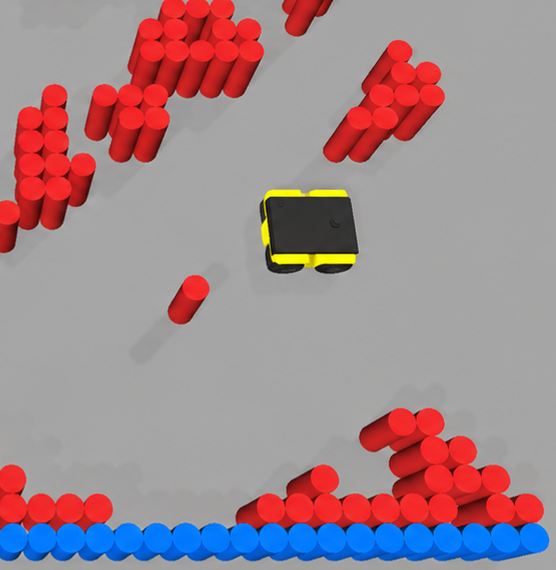}
\par\small (c) BARN: testing
\end{minipage}
\captionof{figure}{Simulation environments used for policy training and evaluation.}
\label{fig:navigation_maps}
\par\vspace{1.2ex}
{%
\centering
\captionof{table}{Robot Footprints and Sensors for Physical Deployment and BARN Simulation}
\label{tab:physical_robot_configurations}
\small
\setlength{\tabcolsep}{3.5pt}
\renewcommand{\arraystretch}{1.2}
\begin{tabular*}{\linewidth}{@{\extracolsep{\fill}}>{\centering\arraybackslash}m{0.25\linewidth}>{\centering\arraybackslash}m{0.32\linewidth}>{\centering\arraybackslash}m{0.37\linewidth}@{}}
\toprule
Robot & \shortstack[c]{Footprint\\[-2pt]\rule{6.2em}{0.3pt}\\[-1pt]Dimensions (m)} & Sensor / FOV \\
\midrule
TurtleBot2 & \shortstack[c]{Circle\\[-2pt]\rule{6.2em}{0.3pt}\\[-1pt]Diameter 0.40} & 2D LiDAR / 270$^\circ$ \\
\addlinespace[3pt]
Go2 & \shortstack[c]{Rectangle\\[-2pt]\rule{6.2em}{0.3pt}\\[-1pt]0.80\,\(\times\)\,0.41} & 3D LiDAR / 360$^\circ$ \\
\addlinespace[3pt]
K1 & \shortstack[c]{Rectangle\\[-2pt]\rule{6.2em}{0.3pt}\\[-1pt]0.20\,\(\times\)\,0.40} & Depth camera / 104$^\circ$ \\
\addlinespace[3pt]
F-shaped robot & \shortstack[c]{Nonconvex polygon\\[-2pt]\rule{8.1em}{0.3pt}\\[-1pt]1.00\,\(\times\)\,0.50} & 2D LiDAR / 360$^\circ$ \\
\addlinespace[3pt]
Jackal (BARN) & \shortstack[c]{Rectangle\\[-2pt]\rule{6.2em}{0.3pt}\\[-1pt]0.42\,\(\times\)\,0.21} & 2D LiDAR / 270$^\circ$ \\
\bottomrule
\end{tabular*}
\par\smallskip
\parbox{\linewidth}{\footnotesize Dimensions are length $\times$ width except for TurtleBot2. For the F-shaped robot, they describe the enclosing rectangle.}\par
}
\end{minipage}\hfill
\usebox{\experimentrewardbox}
\par\vspace{1.5ex}
\end{figure*}

\subsection{Large-Batch Off-Policy Training}
\label{sec:fast_training}

After replay warm-up, each training cycle advances $N$ environments for $L$ vector steps and performs $U$ critic updates, with actor and target updates following the configured schedules. Each vector step advances all $N$ environments. A replay batch provides the transitions for all $U$ updates in a cycle. Once this batch is available, the collector starts collection for the next cycle before the current learner updates begin. The collector uses a separate copy of the actor, called the collection policy, for batched inference. Collection and learner updates are scheduled separately on the GPU so that their execution can overlap. Updated actor weights are copied to the collection policy after each training cycle, and the collector waits for the copy to finish before using the new weights.

Each replay batch contains $U \times B$ transitions sampled uniformly from replay and divided into $U$ minibatches of size $B$. Two preallocated batch buffers separate learner updates from preparation of the next replay batch. While the learner uses one buffer, the next replay batch is prepared in the other once new collection data are available. Sampling waits for the required replay writes to finish, and the learner accesses a prepared batch only after sampling is complete. Preparing one replay batch for multiple updates allows sampling to overlap with learner updates. FastSAC~\cite{seo2025fastsac}, FastTD3~\cite{seo2025fasttd3}, and FastDSAC~\cite{xue2026fastdsac} share this collection and replay interface, with learner-specific update schedules.

\subsection{Training with Arbitrary Robot Footprints}
\label{sec:arbitrary_footprints}

Training navigation policies for robots with nonconvex footprints requires collision checks that follow the occupied region. A single rectangular bounding box covers free space within the recesses of an F-shaped footprint, while the minimum LiDAR range is measured from the sensor origin. FlashNav uses the footprint $\mathcal F$ in the robot specification for collision checking and clearance observations. Simple polygons describe nonconvex outlines, and unions of polygons represent composite footprints.

The footprint interior and boundary are sampled at a configured spatial resolution, and the resulting point set $\mathcal P_{\mathcal F}$ is cached for repeated collision queries. With these points expressed in the body frame, the collision flag at time step $t$ is computed as
\begin{equation}
\label{eq:footprint_collision}
c_t = \bigvee_{\mathbf s\in\mathcal P_{\mathcal F}}
\mathcal M\!\left(R(\psi_t)\mathbf s+\mathbf p_t\right),
\end{equation}
where $\mathbf p_t=(x_t,y_t)$, $R(\psi_t)$ is the planar rotation matrix, and $\mathcal M$ returns the binary occupancy of the map cell containing the queried point. A collision is detected if any footprint sample lies in an occupied cell or outside the map. Coordinate transforms and occupancy queries are batched across environments. This check follows the sampled nonconvex region, with spatial discretization determined by the map resolution and footprint sampling interval.

Following the robot-centered view of collision evaluation~\cite{geng2023robocentric}, we use the footprint $\mathcal F$ to compute clearance features for the policy. Each valid range endpoint $\mathbf e_{t,j}$, where $j$ indexes the beam, is expressed in the body frame using the sensor placement in the robot specification. Its nonnegative distance to the occupied footprint is
\begin{equation}
\label{eq:footprint_clearance}
\delta_{t,j}=\min_{\mathbf f\in\mathcal F}
\left\|\mathbf e_{t,j}-\mathbf f\right\|_2.
\end{equation}

For polygonal footprints, this distance is computed directly from the boundary segments and is set to zero for points inside the occupied region. For a union of polygons, the minimum distance over its components is used. Distances are capped at the sensor range and pooled by taking the minimum within each angular sector. Beams without a valid hit are assigned the maximum range before pooling.

\input{tables/training_time}

\section{Experiments}
\label{sec:experiments}
\subsection{Experimental Setup}
\label{sec:experimental_setup}

Policies were trained on SEnv1 in Fig.~\ref{fig:navigation_maps}(a) using five robot configurations: TurtleBot2, Unitree Go2 (Go2), Booster K1 (K1), an F-shaped robot, and Jackal. Table~\ref{tab:physical_robot_configurations} summarizes the footprints, dimensions, and sensors of the four physical platforms and the Jackal used in BARN simulation. K1 used omnidirectional commands; the other four configurations used differential-drive commands.

FastDSAC, FastSAC, and FastTD3 were compared on three hosts: an AMD Ryzen 9 9950X with an RTX 5090 (32~GB GPU memory), an Intel Core i9-14900K with an RTX 4090 (24~GB), and an Intel Core Ultra 7 265K with an RTX 5060 Ti (16~GB).
During training, the actor and critic networks each used four hidden layers with 100 units per layer.
Training times, episode returns, training throughput, and SEnv2 navigation success rates were averaged over ten training seeds per configuration. All ten seeds reached both thresholds reported in Table~\ref{tab:training_time}. The training success rate was computed as the fraction of successful episodes among the most recent 100 completed episodes. Training pipeline time was the sum of measured intervals for collection waiting, replay preparation, learner updates, and weight synchronization. Episode returns were computed over the same 100-episode window.

For physical deployment, TurtleBot2 was equipped with a Hokuyo UTM-30LX, while the F-shaped robot used a TurtleBot2 base with an LSLiDAR N10P. Both carried an onboard PC with an Intel Core i9-12900H CPU. Go2 was equipped with a Livox MID360 and an NVIDIA Jetson Orin NX onboard computer. K1 used a stereo depth camera. The 3D LiDAR and depth-camera measurements were converted to 2D laser scans before constructing the policy observations. For TurtleBot2 and the F-shaped robot, the testing environments were mapped using ROS GMapping~\cite{Gmapping}, and robot poses were estimated using ROS AMCL~\cite{AMCL}.

\subsection{Training Efficiency}
\label{sec:training_efficiency}

To quantify the training efficiency of FlashNav across robot configurations and computing platforms, both training time and learning progress were evaluated. Table~\ref{tab:training_time} reports $T_{90}$ and $T_{100}$, the wall-clock times to first reach training success rates of 90\% and 100\%, respectively. Fig.~\ref{fig:reward_time} complements these measurements with episode return curves showing how learning progressed over training pipeline time.

With FastDSAC on RTX 5090, TurtleBot2 first reached a training success rate of 100\% in a mean of 16.8~s. Across all five robot configurations on this GPU, the mean $T_{100}$ of FastDSAC ranged from 16.8 to 32.9~s. FastDSAC was the fastest learner overall, with the lowest mean $T_{90}$ in every robot--GPU combination. Hardware also affected training efficiency: the RTX 5090 and RTX 4090 platforms generally required less time than the RTX 5060 Ti platform to reach the same training success rates.
As shown in Fig.~\ref{fig:reward_time}, FastDSAC generally converged faster and required less time to complete the fixed number of training environment steps than FastSAC and FastTD3. This trend was consistent with the shorter times to reach the target training success rates in Table~\ref{tab:training_time}.

\begin{figure}[t]
\centering
\includegraphics[width=0.9\columnwidth]{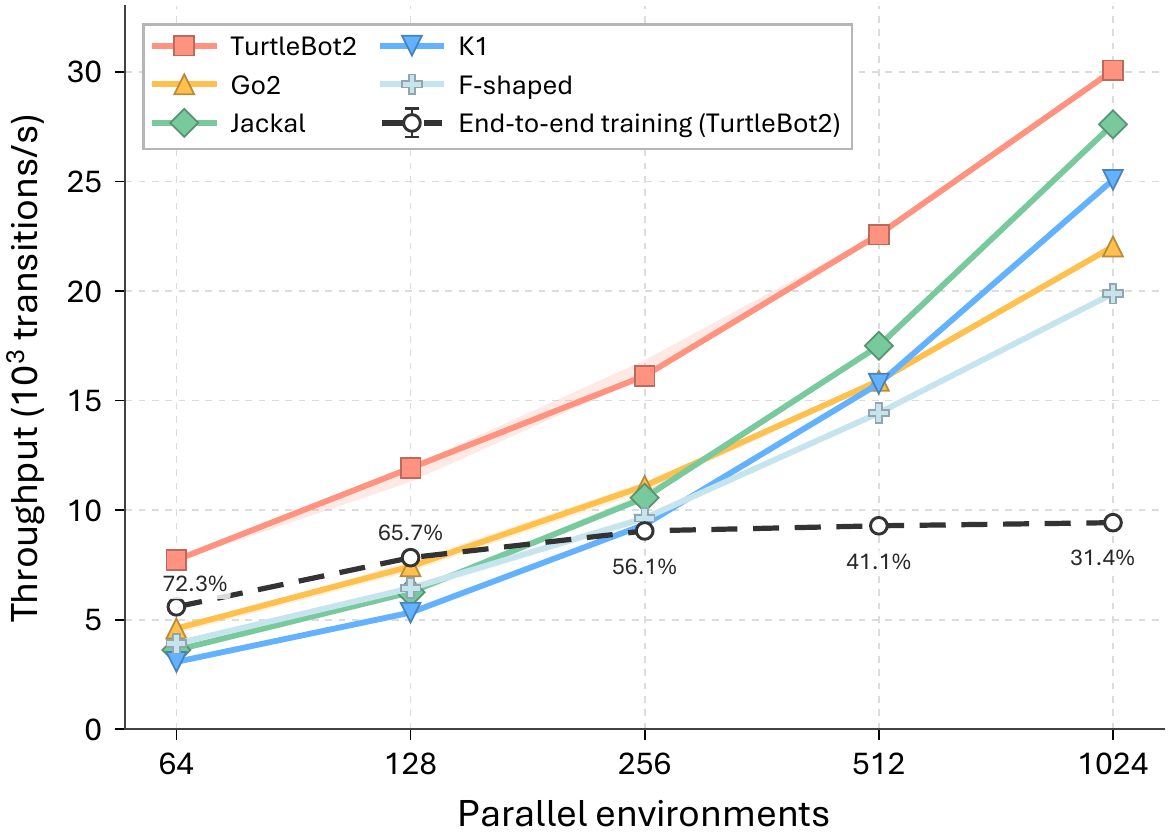}
\caption{Standalone simulator and end-to-end training throughput on RTX 5090. Solid curves show medians and interquartile ranges over five simulator measurements. The dashed curve shows the mean and sample standard deviation over ten TurtleBot2 FastDSAC training seeds; percentage labels give the training-to-simulator throughput ratio.}
\label{fig:simulation_scaling}
\end{figure}

\begin{figure}[t]
\centering
\includegraphics[width=0.98\columnwidth]{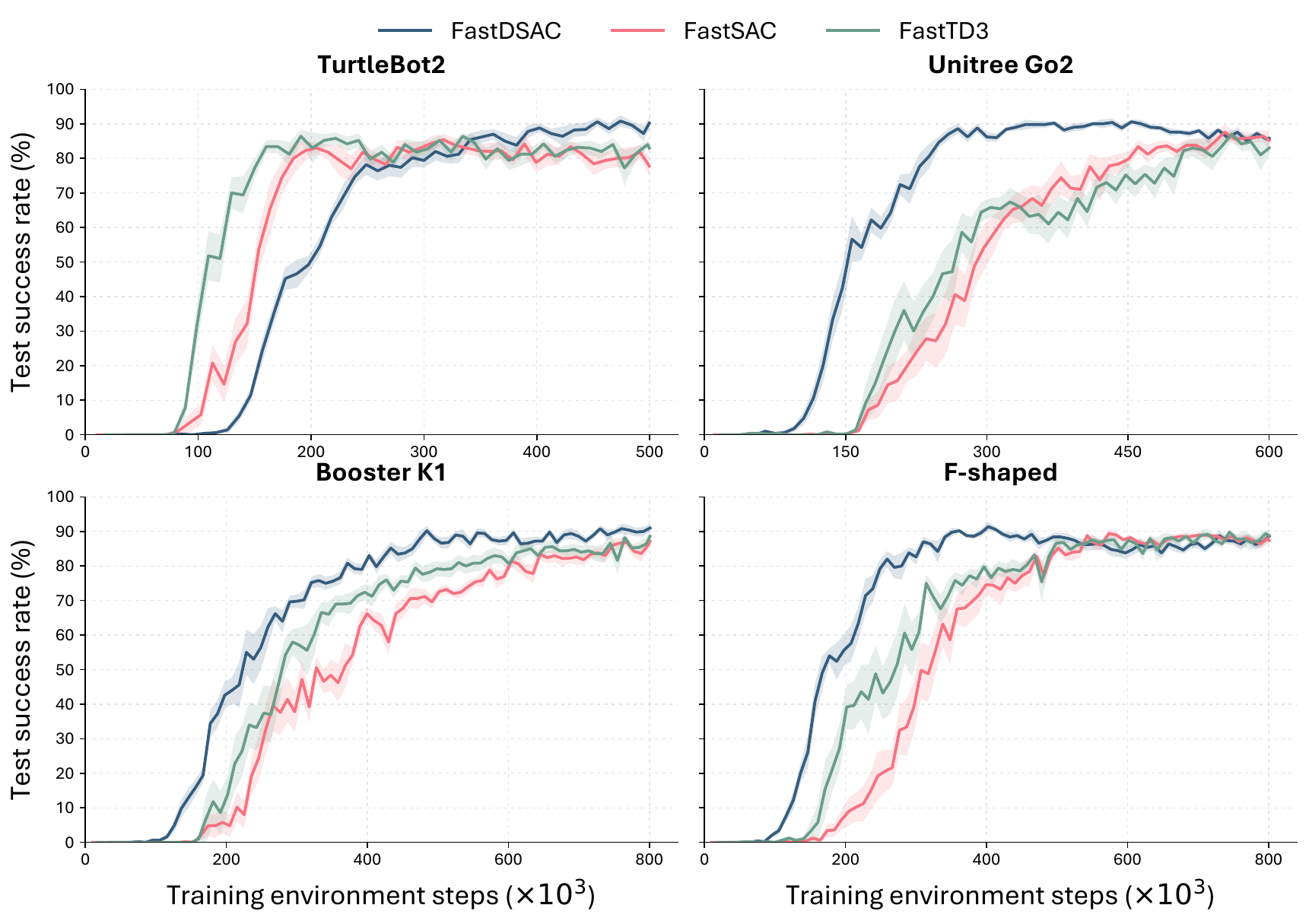}
\caption{Navigation success rates on 50 fixed cases per robot over training environment steps. Curves show means across ten training seeds. Shading indicates one standard error above and below the mean.}
\label{fig:navigation_success}
\end{figure}

\subsection{Parallel Simulation and Training Throughput Scaling}
\label{sec:simulation_scaling}

Standalone simulator throughput and end-to-end training throughput were measured on RTX 5090 with $N=64,128,256,512$, and $1024$. Standalone simulator throughput was measured using precomputed actions, with 100 warm-up vector steps followed by 1000 timed vector steps and GPU synchronization at both timing boundaries. Each robot configuration was measured five times. The dashed black curve reports end-to-end training throughput for TurtleBot2 with FastDSAC over the full process wall time, including environment execution, policy inference, replay operations, and learner updates. The same learner settings and amount of newly collected data per training cycle were used at each $N$.
As shown in Fig.~\ref{fig:simulation_scaling}, increasing $N$ from 64 to 1024 raised standalone simulator throughput by 3.9--8.2 times across the five robot configurations. Over the same range, TurtleBot2 end-to-end training throughput increased by 1.69 times, while its ratio to standalone simulator throughput decreased from 72.3\% to 31.4\%. 

\subsection{Training and Simulator Ablations}
\label{sec:system_ablation}

The ablations assessed how training scheduling and simulator optimizations affected computational efficiency. Both comparisons used TurtleBot2 on RTX 5090 at $N=1024$. The training comparison in Table~\ref{tab:system_ablation} (a) used FastDSAC with identical training budgets. The sequential pipeline used serial collection and learning with per-update replay sampling; the complete pipeline enabled collection--learning overlap and batched replay with double buffering. Training times are reported as means with sample standard deviations. The complete pipeline reduced mean process wall-clock time, including initialization and checkpoint saving, by 7.1\%. All runs reached 100\% training success.

Table~\ref{tab:system_ablation} (b) compares four simulator variants. ``All samples'' evaluates all distance samples along each ray at once; ``Blockwise'' processes them in blocks, skipping rays that have hit obstacles. When episodes end, their environments are reset to start new episodes. Observations are then recomputed for all environments (``Full batch'') or only the reset environments (``Ended only''). All variants used the same map, actions, ray resolution, and collision rules; values are medians over five repetitions. Each optimization increased throughput; together, they yielded a paired median speedup of 2.76 times and reduced peak allocated GPU memory by 80.0\%.

\input{tables/system_ablation}

\subsection{Navigation Performance}
\label{sec:navigation_performance}

Navigation was evaluated across robot configurations in FlashNav simulation and against existing methods in the BARN Challenge~\cite{perille2020benchmarking}.

\paragraph{Navigation Performance in FlashNav Simulation}
The trained policies were evaluated on SEnv2 in Fig.~\ref{fig:navigation_maps}(b), a map unseen during training, using 50 fixed point-to-point navigation tasks per robot. Navigation success rates were computed over these tasks using deterministic actions and the parallel evaluation procedure in Section~\ref{sec:batched_simulation}. Fig.~\ref{fig:navigation_success} plots these rates against training environment steps. For Go2 and the F-shaped robot, navigation success rates increased earlier with FastDSAC than with FastSAC or FastTD3. For TurtleBot2 and K1, FastDSAC achieved higher navigation success rates later in training. The mean navigation success rates reached 90.8\%, 90.6\%, 91.0\%, and 91.4\% for FastDSAC on TurtleBot2, Go2, K1, and the F-shaped robot, respectively.

\begin{figure}[t]
\vspace*{6pt}
\centering
{%
\captionof{table}{Navigation Performance Comparison on BARN dataset.}
\label{tab:barn_results}
\small
\setlength{\tabcolsep}{2.2pt}
\renewcommand{\arraystretch}{1.12}
\begin{tabular}{@{}lcrrrc@{}}
\toprule
Method & Score ($\uparrow$) & SR ($\uparrow$) & CR ($\downarrow$) & TR ($\downarrow$) & Global path \\
\midrule
DWA & 0.212 & 46.0\% & 37.0\% & 17.0\% & $\checkmark$ \\
E-Band & 0.286 & 57.7\% & 9.7\% & 32.6\% & $\checkmark$ \\
E2E & 0.326 & 65.3\% & 33.7\% & 1.0\% & $\times$ \\
KUL+FM & 0.387 & 85.3\% & \textbf{2.0\%} & 12.7\% & $\checkmark$ \\
AIMS & 0.386 & 85.8\% & 3.2\% & 11.0\% & $\checkmark$ \\
DRL-VO & 0.253 & 81.0\% & 19.0\% & \textbf{0.0\%} & $\checkmark$ \\
DRL-DCLP & 0.341 & 71.5\% & 9.0\% & 19.5\% & $\times$ \\
DRL-DCLPv2 & 0.388 & 88.0\% & 6.3\% & 5.7\% & $\checkmark$ \\
FlashNav-DSAC & 0.399 & 80.3\% & 11.0\% & 8.7\% & $\times$ \\
FlashNav-DSACv2 & \textbf{0.462} & \textbf{94.0\%} & 6.0\% & \textbf{0.0\%} & $\checkmark$ \\
\bottomrule
\end{tabular}
}
\end{figure}

\paragraph{Navigation Performance Comparison on BARN}
The Benchmark for Autonomous Robot Navigation (BARN) dataset provides a public benchmark for mobile robot navigation in highly cluttered environments~\cite{perille2020benchmarking}. The policies were tested with a simulated differential-drive Jackal with a maximum linear velocity of 2.0~m/s. Fig.~\ref{fig:navigation_maps}(c) shows a BARN testing scenario. Each configuration was evaluated three times in each of 100 environments, giving 300 trials. FlashNav was compared with classical and state-of-the-art navigation methods, including DWA~\cite{fox1997dynamic}, E-Band~\cite{quinlan1993elastic}, the BARN E2E baseline~\cite{xiao2022barn}, KUL+FM~\cite{xiao2023barn}, AIMS~\cite{xiao2024barn}, DRL-VO~\cite{xie2023drlvo}, and DRL-DCLP~\cite{zhang2025drldclp}.

FlashNav-DSAC and FlashNav-DSACv2 denote two configurations trained with FastDSAC. The former navigates directly to the goal, while the latter follows a global path. Global paths for all methods that used them were generated by Navfn-ROS~\cite{Navfn}. FlashNav-DSACv2 used the first point along the global path at least 2~m from the robot as its local target. To reduce collision risk from command delays at high speeds, both configurations used a speed limit based on obstacle distance, following the approach of DRL-VO~\cite{xie2023drlvo}. The maximum linear velocity was reduced to 1~m/s when the nearest obstacle was within 2.2 meters.

\begin{figure*}[t]
\centering
\includegraphics[width=\textwidth]{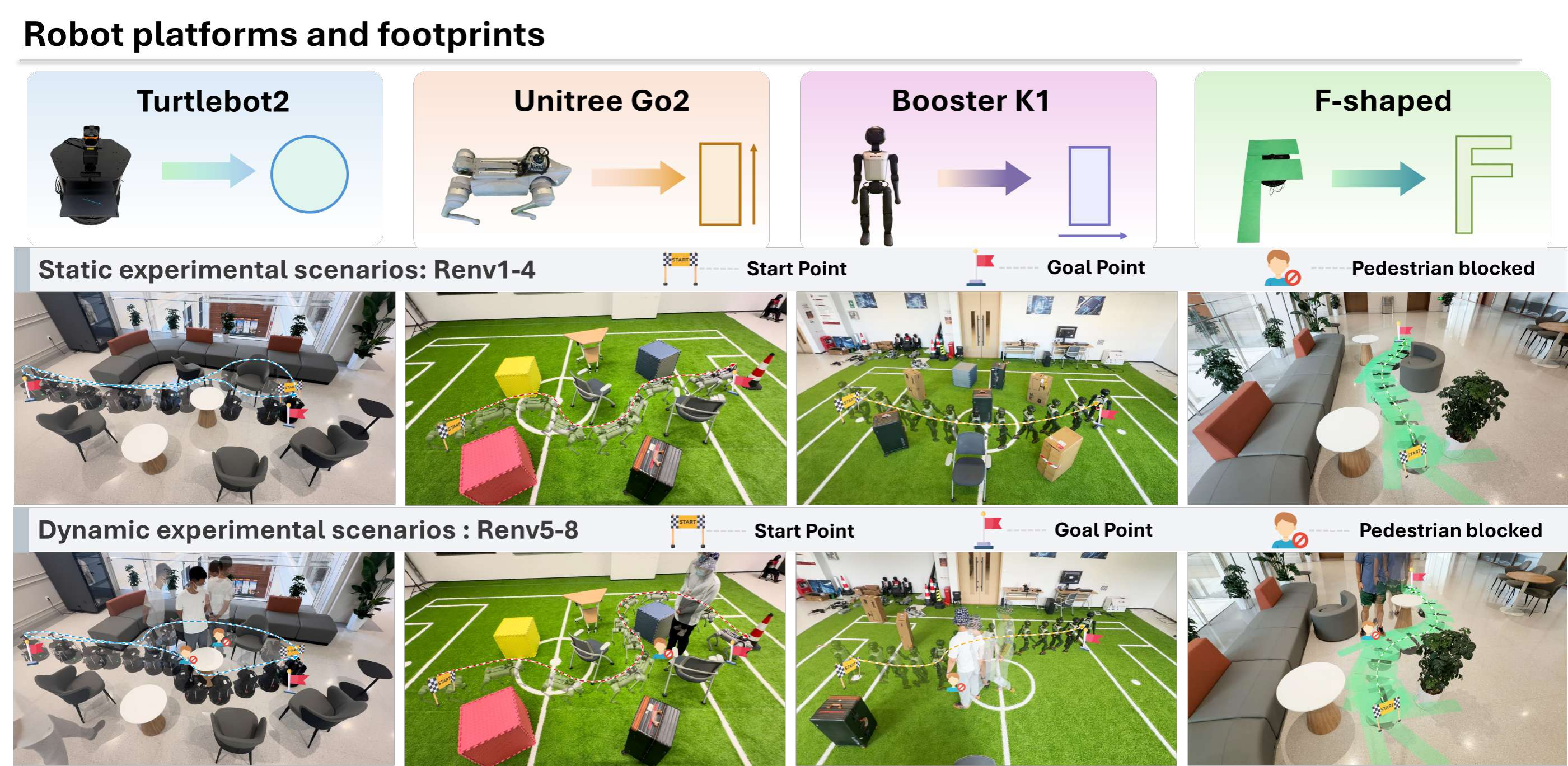}
\caption{Real-world navigation on TurtleBot2, Go2, K1, and the F-shaped robot, from left to right. The top row shows the platforms and footprint representations. The middle and bottom rows show static scenes Renv1--4 and dynamic scenes Renv5--8, respectively. Overlaid robot poses trace the recorded motion, and red flags mark the goals.}
\label{fig:realworld_deployment}
\end{figure*}

The evaluation metrics were navigation success rate (SR), collision rate (CR), timeout rate (TR), and the mean of the per-trial score defined in~\cite{xiao2024barn}
\begin{equation}
\label{eq:barn_score}
s_i=\mathbf{1}_{\mathrm{success},i}
\frac{\mathrm{OT}_i}{\operatorname{clip}(\mathrm{AT}_i,2\mathrm{OT}_i,8\mathrm{OT}_i)},
\end{equation}
where $\mathbf{1}_{\mathrm{success},i}$ equals one if trial $i$ reaches the goal without collision and zero otherwise. Here, $\mathrm{AT}_i$ is the actual navigation time, and $\mathrm{OT}_i$ is the reference time obtained by dividing the benchmark path length by the maximum speed. The clipping operation bounds $\mathrm{AT}_i$ between $2\mathrm{OT}_i$ and $8\mathrm{OT}_i$. Higher scores indicate better performance.

As shown in Table~\ref{tab:barn_results}, FlashNav-DSACv2 outperformed the compared classical and learning-based methods in navigation success rate and score, reaching 94.0\% and 0.462, respectively. The navigation success rate was substantially higher than those of DWA and E-Band and exceeded that of DRL-DCLPv2 by 6.0 percentage points. Without a global path, FlashNav-DSAC also achieved a higher navigation success rate and score than E2E and DRL-DCLP.

\subsection{Real-World Deployment}
\label{sec:realworld_deployment}

Real-world experiments were conducted in indoor scenes arranged to represent offices and other cluttered indoor environments. Static scenes contained tables, seating, boxes, and freestanding obstacles. In dynamic scenes, pedestrians unexpectedly stepped into the robot paths, requiring the robots to respond in real time and avoid them while continuing toward the goals. Fig.~\ref{fig:realworld_deployment} presents static scenes Renv1--4 and dynamic scenes Renv5--8, with one scene of each type for every physical platform. The figure marks the start and goal positions and the locations where pedestrians blocked the paths; overlaid robot poses show the recorded motion.

FastDSAC policies for real-world deployment were trained on RTX 5090. For each robot configuration, the first checkpoint with a navigation success rate of at least 85\% on SEnv2 was selected for real-world deployment. Table~\ref{tab:deployment_checkpoints} shows that the corresponding mean training times were only 29.9--50.2~s across the four robot configurations.
Each robot was evaluated in five repeated trials in its static scene and reached the goal in every trial. In the dynamic scenes, the robots also reached their goals while avoiding pedestrians that blocked their paths.

\input{tables/deployment_checkpoints}

\section{Conclusion and Future Work}
\label{sec:conclusion}

In this work, we present FlashNav, a GPU-based framework for rapidly training deployable navigation policies on a single consumer-grade GPU. Robot specifications configure lightweight simulation for different geometries, sensors, and motion constraints, while batched GPU execution and coordinated off-policy learning shorten training. Ablations quantified the computational benefits of simulator optimizations and training scheduling. The trained policies achieved high navigation success rates in simulation and completed tasks on different physical platforms. Their deployment without additional policy training supports the use of simplified navigation simulation to prepare policies for physical robots. By reducing the time needed to obtain robot-specific policies, FlashNav makes DRL navigation more practical for physical deployment.
Future work will extend training to multi-agent interaction and dynamic environments, and examine a wider range of training maps and learning algorithms.

\bibliographystyle{IEEEtran-no-url}
\bibliography{ref}

\end{document}

%% file: tables/training_time.tex
\begin{table*}[t]
\vspace*{6pt}
\centering
\caption{Mean wall-clock time (s) to first reach 90\% and 100\% training success rate.}
\label{tab:training_time}
\fontsize{9}{10.5}\selectfont
\setlength{\tabcolsep}{2pt}
\renewcommand{\arraystretch}{1.0}
\setlength{\aboverulesep}{0.2ex}
\setlength{\belowrulesep}{0.3ex}
\begin{tabular*}{\textwidth}{@{\extracolsep{\fill}}ll*{9}{r}@{}}
\toprule
\multirow{2}{*}{Robot} & \multirow{2}{*}{Metric} & \multicolumn{3}{c}{RTX 5090} & \multicolumn{3}{c}{RTX 4090} & \multicolumn{3}{c}{RTX 5060 Ti} \\
\cmidrule(lr){3-5}\cmidrule(lr){6-8}\cmidrule(lr){9-11}
 & & FastDSAC & FastSAC & FastTD3 & FastDSAC & FastSAC & FastTD3 & FastDSAC & FastSAC & FastTD3 \\
\midrule
\multirow{2}{*}{TurtleBot2} & $T_{90}$ & \textbf{15.3} & 26.7 & 44.1 & 15.9 & 32.5 & 47.1 & 25.2 & 42.4 & 59.1 \\
 & $T_{100}$ & \textbf{16.8} & 31.7 & 47.9 & 18.5 & 36.3 & 51.4 & 28.5 & 50.8 & 62.3 \\
\midrule
\multirow{2}{*}{Go2} & $T_{90}$ & \textbf{20.3} & 26.9 & 50.6 & 22.1 & 32.5 & 53.9 & 35.3 & 49.5 & 74.5 \\
 & $T_{100}$ & \textbf{21.0} & 28.0 & 60.1 & 23.1 & 34.3 & 67.2 & 36.9 & 56.9 & 87.6 \\
\midrule
\multirow{2}{*}{K1} & $T_{90}$ & \textbf{23.2} & 28.2 & 38.4 & 24.5 & 32.0 & 40.2 & 42.6 & 50.1 & 57.8 \\
 & $T_{100}$ & \textbf{32.9} & 38.0 & 56.4 & 36.5 & 39.5 & 64.5 & 64.4 & 60.3 & 101.4 \\
\midrule
\multirow{2}{*}{F-shaped} & $T_{90}$ & \textbf{23.3} & 34.2 & 52.4 & 25.0 & 39.9 & 59.4 & 40.8 & 56.0 & 73.1 \\
 & $T_{100}$ & \textbf{25.4} & 37.5 & 73.9 & 28.2 & 46.5 & 74.6 & 45.4 & 65.2 & 95.4 \\
\midrule
\multirow{2}{*}{Jackal} & $T_{90}$ & \textbf{14.5} & 21.6 & 32.7 & 14.7 & 26.4 & 35.5 & 22.6 & 32.7 & 44.0 \\
 & $T_{100}$ & 17.3 & 25.3 & 35.7 & \textbf{17.2} & 29.9 & 38.5 & 26.9 & 38.5 & 48.7 \\
\bottomrule
\end{tabular*}
\end{table*}

%% file: tables/system_ablation.tex
\begin{table}[t]
\vspace*{6pt}
\centering
\caption{Training and Simulator Ablations}
\label{tab:system_ablation}
\fontsize{9}{10.5}\selectfont
\setlength{\tabcolsep}{2.2pt}
\renewcommand{\arraystretch}{1.08}
\begin{tabular*}{0.96\columnwidth}{@{\extracolsep{\fill}}
>{\centering\arraybackslash}m{0.28\columnwidth}
>{\centering\arraybackslash}m{0.30\columnwidth}
>{\centering\arraybackslash}m{0.30\columnwidth}@{}}
\multicolumn{3}{c}{(a) Training Pipeline} \\[-1pt]
\toprule
Pipeline & Process time (s) & $T_{100}$ (s) \\
\midrule
Sequential & \mbox{53.73\,$\pm$\,0.45} & \mbox{17.33\,$\pm$\,2.07} \\
Complete & \mbox{49.90\,$\pm$\,1.67} & \mbox{17.03\,$\pm$\,2.55} \\
\bottomrule
\end{tabular*}

\vspace{0.7ex}

\begin{tabular*}{0.96\columnwidth}{@{\extracolsep{\fill}}
>{\centering\arraybackslash}m{0.18\columnwidth}
>{\centering\arraybackslash}m{0.25\columnwidth}
>{\centering\arraybackslash}m{0.24\columnwidth}
>{\centering\arraybackslash}m{0.21\columnwidth}@{}}
\multicolumn{4}{c}{(b) Simulator} \\[-1pt]
\toprule
Ray queries & \shortstack{Post-reset\\recomputation} & \shortstack{Throughput\\($10^3$ trans./s)} & \shortstack{Peak memory\\(GiB)} \\
\midrule
All samples & Full batch & 12.46 & 8.28 \\
Blockwise & Full batch & 22.47 & 1.67 \\
All samples & Ended only & 21.98 & 8.27 \\
Blockwise & Ended only & 34.21 & 1.65 \\
\bottomrule
\end{tabular*}
\end{table}

%% file: tables/deployment_checkpoints.tex
\begin{table}[!t]
\centering
\caption{Training Statistics and Static Real-World Success of Deployed Policies}
\label{tab:deployment_checkpoints}
\fontsize{9}{10.5}\selectfont
\setlength{\tabcolsep}{2.2pt}
\renewcommand{\arraystretch}{1.15}
\begin{tabular*}{0.95\columnwidth}{@{\extracolsep{\fill}}
>{\centering\arraybackslash}m{0.20\columnwidth}
>{\centering\arraybackslash}m{0.23\columnwidth}
>{\centering\arraybackslash}m{0.22\columnwidth}
>{\centering\arraybackslash}m{0.20\columnwidth}@{}}
\toprule
Robot & \shortstack{Training Steps\\($\times 10^4$)} & \shortstack{Training Time\\(s)} & \shortstack{Static\\Success} \\
\midrule
TurtleBot2 & 34.1 & 37.7 & 5/5 \\
Go2 & 25.9 & 29.9 & 5/5 \\
K1 & 43.3 & 50.2 & 5/5 \\
F-shaped & 31.0 & 37.2 & 5/5 \\
\bottomrule
\end{tabular*}
\end{table}

%% file: ref.bib
@article{Zhang2022IPAPRec,
  author = {Zhang, Wei and Zhang, Yunfeng and Liu, Ning and Ren, Kai and Wang, Pengfei},
  title = {{IPAPRec}: A Promising Tool for Learning High-Performance Mapless Navigation Skills With Deep Reinforcement Learning},
  journal = {IEEE/ASME Transactions on Mechatronics},
  volume = {27},
  number = {6},
  pages = {5451--5461},
  year = {2022},
  doi = {10.1109/TMECH.2022.3182427}
}

@article{jia2026unilab,
  title         = {{UniLab}: A Heterogeneous Architecture for Robot {RL} Beyond {GPU}-Dominant Paradigms},
  author        = {Yufei Jia and Zhanxiang Cao and Mingrui Yu and Heng Zhang and Shenyu Chen and Dixuan Jiang and Meng Li and Xiaofan Li and Yiyang Liu and Junzhe Wu and Zheng Li and XiLin Fang and Tingyu Cui and Shengcheng Fu and Haoyang Li and Anqi Wang and Zifan Wang and Dongjie Zhu and Chenyu Cao and Zhenbiao Huang and Ziang Zheng and Jie Lu and Xin Ma and Zhengyang Wei and Xiang Zhao and Tianyue Zhan and Ye He and Yuxiang Chen and Yizhou Jiang and Yue Li and Haizhou Ge and Yuhang Dong and Fan Jia and Ziheng Zhang and Meng Zhang and Xiwa Deng and Zhixing Chen and Hanyang Shao and Chenxin Dong and Yixuan Li and Yizhi Chen and Bokui Chen and Kaifeng Zhang and Hanqing Cui and Yusen Qin and Ruqi Huang and Lei Han and Tiancai Wang and Xiang Li and Yue Gao and Guyue Zhou},
  journal       = {arXiv preprint arXiv:2605.30313},
  year          = {2026},
}

@article{weng2022envpool,
  title={{EnvPool}: A highly parallel reinforcement learning environment execution engine},
  author={Weng, Jiayi and Lin, Min and Huang, Shengyi and Liu, Bo and Makoviichuk, Denys and Makoviychuk, Viktor and Liu, Zichen and Song, Yufan and Luo, Ting and Jiang, Yukun and others},
  journal={Advances in Neural Information Processing Systems},
  volume={35},
  pages={22{}409--22{}421},
  year={2022}
}

@article{xie2023drlvo,
  title={{DRL-VO}: Learning to Navigate Through Crowded Dynamic Scenes Using Velocity Obstacles},
  author={Xie, Zhanteng and Dames, Philip},
  journal={IEEE Transactions on Robotics},
  volume={39},
  number={4},
  pages={2700--2719},
  year={2023},
  doi={10.1109/TRO.2023.3257549}
}

@article{makoviychuk2021isaacgym,
  title={{Isaac Gym}: High Performance {GPU}-Based Physics Simulation for Robot Learning},
  author={Makoviychuk, Viktor and Wawrzyniak, Lukasz and Guo, Yunrong and Lu, Michelle and Storey, Kier and Macklin, Miles and Hoeller, David and Rudin, Nikita and Allshire, Arthur and Handa, Ankur and State, Gavriel},
  journal={arXiv preprint arXiv:2108.10470},
  year={2021}
}

@article{xu2025navrl,
  title={{NavRL}: Learning Safe Flight in Dynamic Environments},
  author={Xu, Zhefan and Han, Xinming and Shen, Haoyu and Jin, Hanyu and Shimada, Kenji},
  journal={IEEE Robotics and Automation Letters},
  volume={10},
  number={4},
  pages={3668--3675},
  year={2025}
}

@article{liu2026height,
  title={{HEIGHT}: Heterogeneous Interaction Graph Transformer for Robot Navigation in Crowded and Constrained Environments},
  author={Liu, Shuijing and Xia, Haochen and Pouria, Fatemeh Cheraghi and Hong, Kaiwen and Chakraborty, Neeloy and Hu, Zichao and Biswas, Joydeep and Driggs-Campbell, Katherine},
  journal={IEEE Transactions on Automation Science and Engineering},
  year={2026},
  doi={10.1109/TASE.2025.3646588}
}

@article{han2025neupan,
  title={{NeuPAN}: Direct Point Robot Navigation With End-to-End Model-Based Learning},
  author={Han, Ruihua and Wang, Shuai and Wang, Shuaijun and Zhang, Zeqing and Chen, Jianjun and Lin, Shijie and Li, Chengyang and Xu, Chengzhong and Eldar, Yonina C. and Hao, Qi and Pan, Jia},
  journal={IEEE Transactions on Robotics},
  volume={41},
  pages={2804--2824},
  year={2025},
  doi={10.1109/TRO.2025.3554252}
}

@article{seo2025fasttd3,
  title={{FastTD3}: Simple, Fast, and Capable Reinforcement Learning for Humanoid Control},
  author={Seo, Younggyo and Sferrazza, Carmelo and Geng, Haoran and Nauman, Michal and Yin, Zhao-Heng and Abbeel, Pieter},
  journal={arXiv preprint arXiv:2505.22642},
  year={2025}
}

@article{seo2025fastsac,
  title={Learning Sim-to-Real Humanoid Locomotion in 15 Minutes},
  author={Seo, Younggyo and Sferrazza, Carmelo and Chen, Juyue and Shi, Guanya and Duan, Rocky and Abbeel, Pieter},
  journal={arXiv preprint arXiv:2512.01996},
  year={2025}
}

@article{xue2026fastdsac,
  title={{FastDSAC}: Unlocking the Potential of Maximum Entropy {RL} in High-Dimensional Humanoid Control},
  author={Xue, Jun and Wang, Junze and Wang, Shanze and Zhang, Xinming and Chen, Yanjun and Zhang, Wei},
  journal={arXiv preprint arXiv:2603.12612},
  year={2026}
}

@misc{Gmapping,
  title = {{gmapping}},
  howpublished = {\url{http://wiki.ros.org/gmapping}},
  url = {http://wiki.ros.org/gmapping},
  note = {Accessed: Sep. 10, 2026}
}

@misc{AMCL,
  title = {{AMCL}},
  howpublished = {\url{http://wiki.ros.org/amcl}},
  url = {http://wiki.ros.org/amcl},
  note = {Accessed: Sep. 10, 2026}
}

@article{wang2025color, title={Train a real-world local path planner in one hour via partially decoupled reinforcement learning and vectorized diversity}, volume={141}, ISSN={0952-1976}, url={http://dx.doi.org/10.1016/j.engappai.2024.109726}, DOI={10.1016/j.engappai.2024.109726}, journal={Engineering Applications of Artificial Intelligence}, publisher={Elsevier BV}, author={Xin, Jinghao and Kim, Jinwoo and Li, Zhi and Li, Ning}, year={2025}, month=Feb, pages={109726} }

@inproceedings{samvelyan2024jaxnav,
 author = {Rutherford, Alex and Beukman, Michael and Willi, Timon and Lacerda, Bruno and Hawes, Nick and Foerster, Jakob},
 booktitle = {Advances in Neural Information Processing Systems},
 doi = {10.52202/079017-0512},
 editor = {A. Globerson and L. Mackey and D. Belgrave and A. Fan and U. Paquet and J. Tomczak and C. Zhang},
 pages = {16{}071--16{}101},
 publisher = {Curran Associates, Inc.},
 title = {No Regrets: Investigating and Improving Regret Approximations for Curriculum Discovery},
 url = {https://proceedings.neurips.cc/paper_files/paper/2024/file/1d0ed12c3fda52f2c241a0cebcf739a6-Paper-Conference.pdf},
 volume = {37},
 year = {2024}
}

@inproceedings{kastner2024murosim, title={{MuRoSim} -- A Fast and Efficient Multi-Robot Simulation for Learning-based Navigation}, url={http://dx.doi.org/10.1109/ICRA57147.2024.10610375}, DOI={10.1109/icra57147.2024.10610375}, booktitle={2024 IEEE International Conference on Robotics and Automation (ICRA)}, publisher={IEEE}, author={Jestel, Christian and R{\"o}sner, Karol and Dietz, Niklas and Bach, Nicolas and E{\ss}er, Julian and Finke, Jan and Urbann, Oliver}, year={2024}, month=May, pages={16881--16887} }

@inproceedings{perille2020benchmarking, title={Benchmarking Metric Ground Navigation}, url={http://dx.doi.org/10.1109/SSRR50563.2020.9292572}, DOI={10.1109/ssrr50563.2020.9292572}, booktitle={2020 IEEE International Symposium on Safety, Security, and Rescue Robotics (SSRR)}, publisher={IEEE}, author={Perille, Daniel and Truong, Abigail and Xiao, Xuesu and Stone, Peter}, year={2020}, month=Nov, pages={116--121} }

@article{zhang2025drldclp,
  title = {{DRL-DCLP}: A Deep Reinforcement Learning-Based Dimension-Configurable Local Planner for Robot Navigation},
  author = {Zhang, Wei and Wang, Shanze and Tan, Mingao and Yang, Zhibo and Wang, Xianghui and Shen, Xiaoyu},
  journal = {IEEE Robotics and Automation Letters},
  volume = {10},
  number = {4},
  pages = {3636--3643},
  year = {2025},
  publisher = {Institute of Electrical and Electronics Engineers (IEEE)},
  doi = {10.1109/lra.2025.3544927},
  url = {https://doi.org/10.1109/LRA.2025.3544927}
}

@article{shacklett2021large,
  title = {Large batch simulation for deep reinforcement learning},
  author = {Shacklett, Brennan and Wijmans, Erik and Petrenko, Aleksei and Savva, Manolis and Batra, Dhruv and Koltun, Vladlen and Fatahalian, Kayvon},
  journal = {arXiv preprint arXiv:2103.07013},
  year = {2021}
}

@inproceedings{li2023pql,
  title = {Parallel $Q$-Learning: Scaling Off-policy Reinforcement Learning under Massively Parallel Simulation},
  author = {Li, Zechu and Chen, Tao and Hong, Zhang-Wei and Ajay, Anurag and Agrawal, Pulkit},
  booktitle = {Proceedings of the 40th International Conference on Machine Learning},
  volume = {202},
  pages = {19{}440--19{}459},
  year = {2023},
  publisher = {PMLR},
  series = {Proceedings of Machine Learning Research},
  url = {https://proceedings.mlr.press/v202/li23f.html}
}

@article{eschmann2024fly,
  title = {Learning to Fly in Seconds},
  author = {Eschmann, Jonas and Albani, Dario and Loianno, Giuseppe},
  journal = {IEEE Robotics and Automation Letters},
  volume = {9},
  number = {7},
  pages = {6336--6343},
  year = {2024},
  publisher = {Institute of Electrical and Electronics Engineers (IEEE)},
  doi = {10.1109/lra.2024.3396025},
  url = {https://doi.org/10.1109/LRA.2024.3396025}
}

@inproceedings{geng2023robocentric,
  title = {Robo-Centric {ESDF}: A Fast and Accurate Whole-Body Collision Evaluation Tool for Any-Shape Robotic Planning},
  author = {Geng, Shuang and Wang, Qianhao and Xie, Lei and Xu, Chao and Cao, Yanjun and Gao, Fei},
  booktitle = {2023 IEEE/RSJ International Conference on Intelligent Robots and Systems (IROS)},
  publisher = {IEEE},
  year = {2023},
  month = oct,
  pages = {290--297},
  doi = {10.1109/IROS55552.2023.10342074},
  url = {https://doi.org/10.1109/IROS55552.2023.10342074}
}

@misc{han2026irsim,
  title = {{IR-SIM}: A Lightweight Skill-Native Simulator for Navigation, Learning, and Benchmarking},
  author = {Han, Ruihua and Wang, Shuai and Li, Chengyang and Gao, Rui and Wang, Xinyi and Liu, Zhe and Li, Guoliang and Lu, Yupu and Hao, Qi and Pan, Jia and Zhao, Hengshuang},
  year = {2026},
  publisher = {arXiv},
  doi = {10.48550/ARXIV.2606.08729},
  url = {https://arxiv.org/abs/2606.08729},
  note = {arXiv:2606.08729}
}

@inproceedings{bettini2022vmas,
  title = {Vmas: A vectorized multi-agent simulator for collective robot learning},
  author = {Bettini, Matteo and Kortvelesy, Ryan and Blumenkamp, Jan and Prorok, Amanda},
  booktitle = {International symposium on distributed autonomous robotic systems},
  pages = {42--56},
  year = {2022},
  organization = {Springer}
}

@article{lan2022warpdrive,
  author  = {Tian Lan and Sunil Srinivasa and Huan Wang and Stephan Zheng},
  title   = {{WarpDrive}: Fast End-to-End Deep Multi-Agent Reinforcement Learning on a {GPU}},
  journal = {Journal of Machine Learning Research},
  year    = {2022},
  volume  = {23},
  number  = {316},
  pages   = {1--6},
  url     = {http://jmlr.org/papers/v23/22-0185.html}
}

@article{wang2026dclppp,
  title = {{DCLP++}: Learning to Navigate with Footprint Clearance and Relative Motion},
  author = {Wang, Shanze and Zhang, Wei},
  journal = {arXiv preprint arXiv:2609.08711},
  year = {2026},
  url = {https://arxiv.org/abs/2609.08711}
}

@article{fox1997dynamic,
  title = {The dynamic window approach to collision avoidance},
  author = {Fox, D. and Burgard, W. and Thrun, S.},
  journal = {IEEE Robotics \& Automation Magazine},
  volume = {4},
  number = {1},
  pages = {23--33},
  year = {1997},
  doi = {10.1109/100.580977}
}

@inproceedings{quinlan1993elastic,
  title = {Elastic bands: connecting path planning and control},
  author = {Quinlan, S. and Khatib, O.},
  booktitle = {[1993] Proceedings IEEE International Conference on Robotics and Automation},
  pages = {802--807},
  year = {1993},
  doi = {10.1109/robot.1993.291936}
}

@article{xiao2022barn,
  title = {Autonomous Ground Navigation in Highly Constrained Spaces: Lessons Learned From the Benchmark Autonomous Robot Navigation Challenge at {ICRA} 2022 [Competitions]},
  author = {Xiao, Xuesu and Xu, Zifan and Wang, Zizhao and Song, Yunlong and Warnell, Garrett and Stone, Peter and Zhang, Tingnan and Ravi, Shravan and Wang, Gary and Karnan, Haresh and Biswas, Joydeep and Mohammad, Nicholas and Bramblett, Lauren and Peddi, Rahul and Bezzo, Nicola and Xie, Zhanteng and Dames, Philip},
  journal = {IEEE Robotics \& Automation Magazine},
  volume = {29},
  number = {4},
  pages = {148--156},
  year = {2022},
  doi = {10.1109/mra.2022.3213466}
}

@article{xiao2023barn,
  title = {Autonomous Ground Navigation in Highly Constrained Spaces: Lessons Learned From the Second {BARN} Challenge at {ICRA} 2023 [Competitions]},
  author = {Xiao, Xuesu and Xu, Zifan and Warnell, Garrett and Stone, Peter and Guinjoan, Ferran Gebelli and Rodrigues, R{\^o}mulo T. and Bruyninckx, Herman and Mandala, Hanjaya and Christmann, Guilherme and Blanco-Claraco, Jose Luis and Rai, Shravan Somashekara},
  journal = {IEEE Robotics \& Automation Magazine},
  volume = {30},
  number = {4},
  pages = {91--97},
  year = {2023},
  doi = {10.1109/mra.2023.3322920}
}

@article{xiao2024barn,
  title = {Autonomous Ground Navigation in Highly Constrained Spaces: Lessons Learned From the Third {BARN} Challenge at {ICRA} 2024 [Competitions]},
  author = {Xiao, Xuesu and Xu, Zifan and Datar, Aniket and Warnell, Garrett and Stone, Peter and Damanik, Joshua Julian and Jung, Jaewon and Deresa, Chala Adane and Huy, Than Duc and Jinyu, Chen and Yichen, Chen and Cahyono, Joshua Adrian and Wu, Jingda and Mo, Longfei and Lv, Mingyang and Lan, Bowen and Meng, Qingyang and Tao, Weizhi and Cheng, Li},
  journal = {IEEE Robotics \& Automation Magazine},
  volume = {31},
  number = {3},
  pages = {197--204},
  year = {2024},
  doi = {10.1109/mra.2024.3427891}
}

@misc{Navfn,
  title = {{navfn}},
  howpublished = {\url{https://wiki.ros.org/navfn}},
  url = {https://wiki.ros.org/navfn},
  note = {Accessed: Sep. 10, 2026}
}

@inproceedings{akmandor2022deep,
  title = {Deep Reinforcement Learning based Robot Navigation in Dynamic Environments using Occupancy Values of Motion Primitives},
  author = {Akmandor, Ne{\c{s}}et {\"U}nver and Li, Hongyu and Lvov, Gary and Dusel, Eric and Padir, Ta{\c{s}}kin},
  booktitle = {2022 IEEE/RSJ International Conference on Intelligent Robots and Systems (IROS)},
  pages = {11{}687--11{}694},
  year = {2022},
  organization = {IEEE},
  doi = {10.1109/IROS47612.2022.9982133}
}

@article{pfeiffer2018reinforced,
  title = {Reinforced Imitation: Sample Efficient Deep Reinforcement Learning for Mapless Navigation by Leveraging Prior Demonstrations},
  author = {Pfeiffer, Mark and Shukla, Samarth and Turchetta, Matteo and Cadena, Cesar and Krause, Andreas and Siegwart, Roland and Nieto, Juan},
  journal = {IEEE Robotics and Automation Letters},
  volume = {3},
  number = {4},
  pages = {4423--4430},
  year = {2018},
  doi = {10.1109/LRA.2018.2869644}
}

@inproceedings{truong2023rethinking,
  title = {Rethinking {Sim2Real}: Lower Fidelity Simulation Leads to Higher {Sim2Real} Transfer in Navigation},
  author = {Truong, Joanne and Rudolph, Max and Yokoyama, Naoki Harrison and Chernova, Sonia and Batra, Dhruv and Rai, Akshara},
  booktitle = {Proceedings of The 6th Conference on Robot Learning},
  pages = {859--870},
  year = {2023},
  volume = {205},
  series = {Proceedings of Machine Learning Research},
  publisher = {PMLR},
  url = {https://proceedings.mlr.press/v205/truong23a.html}
}
